\documentclass[a4paper,UKenglish,cleveref, autoref, thm-restate]{lipics-v2021}

\usepackage{times}  
\usepackage{helvet}  
\usepackage{courier}  
\usepackage{url}  
\usepackage{graphicx}  

\usepackage{amsmath}
\usepackage{amsfonts}
\usepackage{amssymb}

\usepackage{latexsym}
\usepackage{enumerate}
\usepackage{mathtools}
\usepackage{xspace}
\usepackage{soul}
\usepackage{todonotes}
\setuldepth{x} 
\usepackage{algpseudocode}
\usepackage{tikz}
\usepackage{pgf}
\usepackage{pgfplots}
\usepackage{longtable}
\usepackage{array}
\usepackage{colortbl}
\usepackage{booktabs} 
\usepackage{siunitx} 
\usepackage{pgfplotstable} 
\usetikzlibrary{arrows,decorations,backgrounds,automata,shapes,patterns,decorations,backgrounds,plotmarks,calc,fit}
\usepackage{multirow}

\usepackage{cite}

\title{Lazy Probabilistic Roadmaps Revisited} 

\titlerunning{Lazy Probabilistic Roadmaps Revisited} 

\author{Miquel Ramirez}{School of Electric and Electronic Engineering, University of Melbourne, Australia}{miquel.ramirez@unimelb.edu.au}{https://orcid.org/0000-0002-1838-7982}{}
\author{Daniel Selvaratnam}{Division of Decision and Control Systems, School of Electrical Engineering and Computer Science, KTH Royal Institute of Technology, Stockholm}{selv@kth.se}{https://orcid.org/0000-0003-3329-436X}{}
\author{Chris Manzie}{School of Electric and Electronic Engineering, University of Melbourne, Australia}{manziec@unimelb.edu.au}{}{}

\authorrunning{M. Ramirez, D. Selvaratnam, C. Manzie}
\Copyright{The Authors}



\relatedversion{} 

\nolinenumbers 

\hideLIPIcs  

\newcommand{\img}[1]{\mathrm{img}(#1)}
\newcommand{\suchthat}{\mathrm{such}~\mathrm{that}}

\begin{document}

\maketitle

\begin{abstract}
This paper describes a revision of the classic Lazy Probabilistic Roadmaps
algorithm (Lazy PRM), that results from pairing PRM and a novel Branch-and-Cut 
(BC) algorithm. Cuts are dynamically generated constraints that are imposed on 
minimum cost paths over the geometric graphs selected by PRM. Cuts
eliminate paths that cannot be mapped into smooth plans that satisfy 
suitably defined kinematic constraints. We generate candidate smooth plans by
fitting splines to vertices in minimum-cost path. Plans are 
validated with a recently proposed algorithm that maps them into finite traces,
without need to choose a fixed discretization step.
Trace elements exactly describe when plans cross constraint boundaries modulo
arithmetic precision.
We evaluate several planners using our methods over the recently proposed
BARN benchmark, and we report evidence of the scalability of our approach.
\end{abstract}

\section{Introduction}

The problem of planning the motions of a moving object like a robot is, in its most 
basic form, that of finding trajectories that satisfy \emph{constraints}. The classic 
example of constraint in robot motion planning is that of \emph{obstacles}, objects like walls, etc., that we want our controlled robot to avoid being in
contact with or too close by. Besides ensuring that trajectories and constraints
are disjoint, we also want the former to minimize some given index of performance,
that ranks trajectories according to their desirability.

In this paper we look at motion planning as an infinite optimization problem~\cite{hettich:sip},
that is laid upon a given set $C \subset \mathbb{R}^d$, the robot configuration space.
Informally, solutions (\emph{plans})
must stay within the subset $F \coloneqq C \setminus O$, where $O$ represents the obstacles. 
Plans are piece-wise defined curves $\rho: [0,1] \to C$, that connect a given initial configuration 
$q_0 \in F$ and target set $F_{\star} \subset F$. Plans must also
belong to a set $\Pi$ given by suitably defined \emph{kinematic and dynamic} constraints~\cite{canny:94:jacm}. 
The task at hand is to (1) select plans that minimize a smooth \emph{objective function}, 
and (2) verify whether  the image of $\rho$ is contained in $F$, i.e. $\img{\rho} \subset F$. In 
this paper we require plans to be $C^2$-continuous, so their first and second derivatives exist. This enables the assumption
that a suitably designed local controller (see~\cite{ames:19:cbf} for an expressive and recent framework)
that can steer 
the robot within a small neighborhood of $\rho$ without
overlapping with any obstacles. This assumption holds, for example, when the robot dynamics are given by
a differentially flat non-linear dynamical system~\cite{vannieuwstadt:98:traj_gen}.

\begin{figure}[h]
\centering
\includegraphics[width=0.5\textheight]{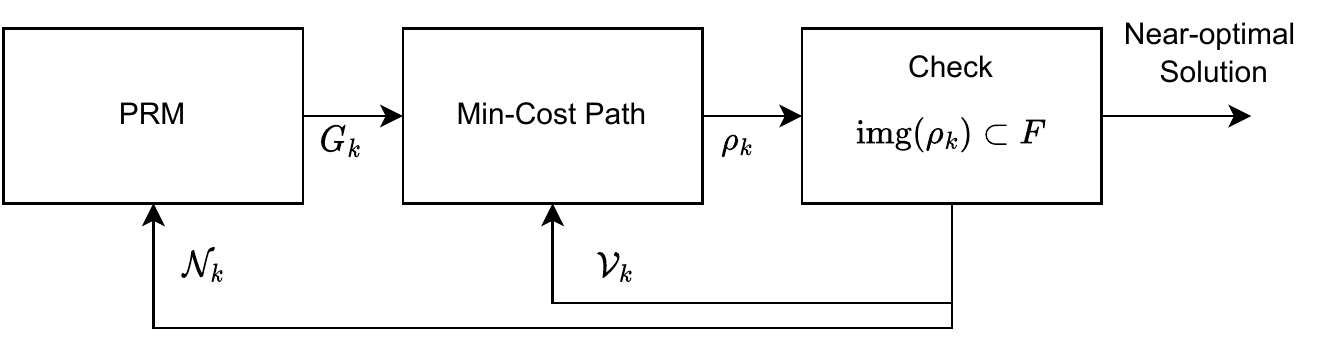}
\caption{Bohlin and Kavraki's decomposition of motion planning. ${\cal N}_k$ and ${\cal V}_k$
	are, respectively, search directions and no-good constraints generated when the incumbent plan $\rho_k$ is found
	to overlap obstacles.}
\label{fig:bk_decomposition}
\end{figure}

The complexity of motion planning motivates its decomposition into more tractable subproblems, 
all the while striking compromises on the optimality of solutions and completeness of the 
resulting algorithm. A concrete and classic example of such a decomposition is
Bohlin and Kavraki's Lazy Probabilistic Roadmaps (Lazy PRM)~\cite{bohlin:lazy_prm:00}. Figure~\ref{fig:bk_decomposition}
depicts Bohlin and Kavraki's iterative algorithm, and the order in which
subproblems are solved. For the $k$th iteration, the first subproblem selects a 
\emph{geometric graph}~\cite{penrose:rgg} $G_k = (V_k, E_k)$. Vertices $V_k \in C$ (points) and edges $E_k$ 
(point-to-point curves) in $G_k$ identify
a finite subset of decision variables of the infinite motion planning problem. The graph
$G_k$ is then used to formulate a \emph{minimum-cost path problem}\cite{korte:combinatorial}, where optimal solutions
correspond to paths (piecewise functions) $\rho_k$ in $G_k$ that minimize a suitably defined cost function 
and connect $q_0$ with $F_{\star}$. If such a $\rho_k$ exists, it is necessary to check whether
$\img{\rho_k} \subset F$. This check fails whenever $\rho_k$ overlaps with some
constraint $O_i$. In this case the algorithm backtracks, and two new constraints are generated\cite{bruynooghe:backtracking,hooker:10:integrated} 
for the roadmap and minimum-cost path search.
For the former, a neighbourhood ${\cal N}_k \subset C$ is defined, that \emph{directs} PRM towards geometric graphs with
vertex set $V \supset V_k$ that may support edges and paths around constraint $O_i$. For the minimum-cost path problem, a 
\emph{no-good} constraint ${\cal V}_k$ $\subset$ $E_k$ is generated
that removes all solutions that would violate constraint $O_i$.
A supergraph $G_{k+1}$ of $G_k$ is then selected, and the procedure repeats until a feasible path is found. 
This algorithm can be shown to be both probabilistically complete (PC) and asymptotically optimal (AO) in a trivial manner from
the results by Karaman and Frazzoli~\cite{karaman:11:ijrr} and later follow-ups~\cite{solovey:20:icra}.

\subsection{Contributions}

This paper expands Bohlin and Kavraki's Lazy PRM algorithm by (1) doing away with the need to sample $\rho$ at a fixed
rate to verify whether $\img{\rho}$ $\subset$ $F$ (section~\ref{efficient_and_exact}), 
(2) solving a minimum-cost path problem augmented with dynamically generated constraints via branch-and-cut 
(section~\ref{section:constrained_best_path}), and (3) enforcing kinematic constraints by means of fitting a $C^2$ curve to waypoints (section~\ref{section:trajectory_fitting}) and checking if constraints on $\dot\rho$ are violated
(section~\ref{velocity_and_kinematic_constraints}).

The problem of establishing whether $\img{\rho} \cap O \neq \emptyset$ can only be solved~\cite{tarski:51} when
$\rho$ and $O$ are described by a system of polynomial
equations. Testing emptiness is then reduced to root-finding.
However, there is no analytical solution
to polynomial root-finding when the degree is greater than $5$~\cite[Abel-Ruffini Theorem]{dummit:AbstractAlgebra2004}. 
We  follow an alternative strategy~\cite{selvaratnam:polynomial_verification}: instead
of locating roots, we seek a sequence of  
\emph{root isolating intervals}~\cite{melhorn:vca}
enclosing them.

Additionally, we generalize Lazy PRM by noting that it is
an application of the \emph{branch-and-cut} (BC) method for combinatorial 
optimization~\cite{wolsey:ip,mitchell:branch_and_cut,hooker:10:integrated}. BC algorithms
consist of two components: \emph{branch-and-bound} (BB) search and a suitably defined
\emph{cutting plane} procedure. The latter produces \emph{cuts}, linear constraints,
that prune away solutions that violate problem constraints that cannot be represented explicitly.
In the context of Lazy PRM,
cuts rule out sequences of edges in the geometric graph that violate kinematic constraints,
or are shown to not satisfy kinematic and geometric constraints simultaneously.

\subsection{Structure of the Paper}
The paper is structured as follows. We start by introducing a number of definitions we rely upon. We then introduce our extension to Bohlin and Kavraki's Lazy PRM, detailing the ways in which we depart from Figure~\ref{fig:bk_decomposition}, the algorithms or optimization problems used.
We then report the results of a comprehensive empirical evaluation over a recently proposed benchmark for motion planning algorithms~\cite{perille:20:benchmarking}. We end the paper covering key related and future work.

\section{Mathematical Framework}
\label{mathematical_framework}

We provide now several basic definitions that will be used throughout the paper. Let $\mathbb{Q}$ be the set of rational numbers, and $\mathbb{R}$ the reals. 
Given two \emph{points}
$x, y \in \mathbb{R}^d$, we denote by $\Vert x - y \Vert_2$ the $l_2$-norm, or \emph{Euclidean distance} between them. Let
$$ X \oplus Y := \{x + y \mid x \in X, y \in Y \}$$
be the \emph{Minkowski sum} of $X \subseteq \mathbb{R}^d$ and $Y \subseteq \mathbb{R}^d$. 
${\cal B}_r(x) := \{ y \in \mathbb{R}^d \mid \|y - x\|_2 \leq r \}$
denotes the $d$-dimensional \emph{closed ball} of radius $r \geq 0$ centered at $x \in \mathbb{R}^d$. Similarly,
given a function $\pi: [0, 1] \to \mathbb{R}^d$, we define 
${\cal B}_{r}(\pi) = \bigcup_{t \in [0, 1]} {\cal B}_{r}(\pi(t))$.  
Given a subset $C \subset \mathbb{R}^d$
we denote by $vol(C)$ its Lebesgue measure or \emph{volume}. 
Given a set of points $X \subset C$,
then $D(X, C)$ is the $\ell_2$-\emph{dispersion} of $X$
\begin{align}
\label{l2:dispersion}
D(X, C) &\coloneqq \sup_{q \in C} \min_{q' \in X} \Vert q - q' \Vert_2  \\
& = \sup \{ r > 0: \exists q \in C\,\mbox{s.t.}\, B_2(q, r) \cap X = \emptyset\}; \nonumber
\end{align}
in words, the radius $r$ of the largest ball ${\cal B}_r(y)$ we can embed within $C\setminus X$ around $y \not\in X$.

Let $R[x_1,...,x_d]$ denote the ring of polynomials in the variables $x_1,...,x_d$ with coefficients in $R$, where $R$ 
can be any commutative ring with multiplicative identity \cite[Chapter 9]{dummit:AbstractAlgebra2004}. Thus, $\mathbb{R}[t]$ 
is the ring of \emph{univariate polynomials} with real-coefficients, and $\mathbb{Q}[x_1,...,x_d]$ the ring of \emph{multivariate polynomials} 
in $d$ variables with rational coefficients. 

A \emph{graph} $G=(V,E)$ is given by its vertex $V$ and edge $E \subseteq V \times V$ sets, with $v \neq w$ for every edge $(v, w) \in E$. If $(v,w) \in E$, we say that $v$ and $w$ are \emph{adjacent}, and that $(v,w)$ is an \emph{outgoing} (resp. \emph{incoming}) edge for $v$ (resp. $w$). A \emph{path} in $G$ is a sequence
$p=v_0,v_1,\ldots,v_n$ of \emph{distinct elements} of $V$ such that $(v_i, v_{i+1}) \in E$ for all $0 \leq i < n$. The set of all
such sequences is $Paths(G)$. The graph $G$ is \emph{connected} if for any two distinct
vertices $u, v \in V$ there is a path $p=w_0, \ldots, w_n$ such that $u=w_0$ and $v=w_n$. 

\subsection{Motion Planning}
\label{motion_planning}

We now fully formalize and describe the discussion of motion planning problems in the Introduction. 
Let the open set $C \subset \mathbb{R}^d$ be the \emph{robot configuration space}, that is, the set of $d$-dimensional
real vectors representing robot states (pose, battery level, etc.), and let $O \subset C$ be a closed subset of
$C$ that correspond to sets of potential robot states (overlap with obstacles, minimum battery levels, etc.) to be avoided. 
$O$ consists of $m$ connected components $O_i$ s.t. $O = \bigcup_{i = 1}^{m} O_i$, and each $O_i$ is a
\emph{semi-algebraic set} of the form
\begin{align}
O_i \coloneqq \bigcap_{j=1}^{m_i}\{ x \in C \,:\, g^{i}_{j}(x) \leq 0\}. \label{eq:Oi}
\end{align}
In words, each obstacle $O_i$ is given by the intersection of $m_i$ sets, each
defined in terms of \emph{constraints} $g^{i}_{j}(x) \leq 0$, where each $g^{i}_{j}: \mathbb{R}^d \to \mathbb{R}$ is 
a multivariate polynomial. The set of admissible or valid robot states, or \emph{freespace}, is then 
the open set $F = C \setminus O$.

Motion planning is the problem of finding a continuous function $\rho: [0, 1] \to C$, or \emph{plan}, that connects a given
initial state $q_0 \in F$, to a \emph{goal} region $F_{\star} \subset F$ s.t. 
$vol(F_{\star}) > 0$. That is, a plan belonging to a suitably defined \emph{plan domain}, $\rho \in \Pi$, that verifies $\rho(0) = q_0$,
$\rho(1) \in F_{\star}$, and every point $\rho(t)$ lies within $F$. Additionally,
we will require $\rho$ to minimize a given \emph{objective} function $f:\Pi \to \mathbb{R}$, which ranks the plans in $\Pi$. We use the following notation to
describe precisely the set of plans $\rho$ that solve optimally this problem
\begin{subequations}
\label{generic_motion_planning}
\begin{align}
	\min_{\rho \in \Pi} & \; f(\rho) & \\
	&\; \rho(0) = q_0,\, \rho(1) \in F_{\star} & \label{gmp:init_goal}\\
	&\; {\cal B}_{\delta_i}(\rho) \subset C \setminus O_i,\,&  i=1,\ldots,m \label{gmp:freespace_constraint}
\end{align}
\end{subequations}
where $\delta_i \geq 0$, or \emph{clearance}, indicates the minimum distance from a point $\rho(t)$ to the boundary of a set 
$O_i$~\eqref{eq:Oi}\footnote{We note that clearance does not need to be a global parameter.}
\emph{Plan domains} $\Pi$ are defined in terms of $m'$ \emph{differential constraints}~\cite[Chapter 13]{lavalle:06} 
\begin{align}
	\label{plan_domain_def}
	\Pi \coloneqq \{ \rho \,:\, g_i(\rho(t), \dot\rho(t), \ddot\rho(t)) \leq 0,\, i=1,\ldots,m' \}
\end{align}
where $\rho$ is a $C^2$-continuous function and each $g^{i}_j: \mathbb{R}^{3d} \to \mathbb{R}$. Candidate plans $\rho$ are defined as \emph{piecewise functions} in $\Pi$ that follow from the concatenation of a finite
sequence of \emph{actions} $(a_i)_{i=1}^{N}$, $1 \leq i \leq N$, where $a_i: [0,1] \to C$. We observe that for any $\rho$ with 
more than one action, $C^2$-continuity is not trivial, and we discuss in section~\ref{section:trajectory_fitting} how to obtain
efficiently piecewise functions with the desired properties. Finally, we locate points on $\rho(t)$ using 
\begin{align}
\label{point_location}
\rho(t) = a_{\lceil nt \rceil}(nt - \lfloor nt \rfloor)
\end{align}

\begin{example}
A well-studied special case
of Problem~\eqref{generic_motion_planning} is known as \emph{geometric motion planning}
, where $\Pi$ is the set of continuous, \emph{piecewise-defined linear functions}. Thus, each $a_i(t)$ is a vector of \emph{linear} polynomials of the form 
\[
a_i(t) = ( q_{i+1} - q_{i})t + q_{i}, 
\]
where $q_1,...,q_N \in \mathbb{R}^d$ are \emph{waypoints}. The plans $\pi \in \Pi$ are required to have $C^0$-continuity; i.e., $a_i(0) = a_{i+1}(1)$, for $i=1,\ldots,N-1$.
\end{example}

\subsection{Probabilistic Roadmaps and Random Geometric Graphs}
\label{probabilistic_roadmaps}

\begin{figure}[ht!]
\begin{algorithmic}[1]
\Procedure{PRM}{$G_k=(V_k, E_k)$, ${\cal X}$}
\State \emph{choose} $q_k$ from $X_k \in {\cal X}$, $E \leftarrow \emptyset$
\For{$q \in \big({\cal B}_{r_k}(q_k) \cap V_k\big)$}
\State \emph{choose} $\lambda \in \Lambda$ s.t. $\lambda(0)=q_k$ \emph{and} $\lambda(1)=q$
\State $E \leftarrow E \cup \{(q_k, q)\}$
\EndFor
\State \emph{return} $G_{k+1} = (V_k \cup \{ q_k \}, E_k \cup E)$
\EndProcedure
\end{algorithmic}
\caption{Graph search or ``learning'' component of the Probabilistic Roadmaps (PRM) algorithm~\cite{kavraki:94:roadmaps} as an iterative geometric graph optimization algorithm.}
\label{alg:prm}
\end{figure}

The key algorithmic development leading to general, scalable algorithms for motion
planning was the invention of the Probabilistic Roadmaps (PRM) 
algorithm~\cite{kavraki:98:roadmaps,solovey:20:prm}. As noted in
the Introduction, motion planning problems~\eqref{generic_motion_planning} cannot be solved
directly, as it is necessary to identify a \emph{finite subset} of the decision variables and
constraints. PRM decomposes~\eqref{generic_motion_planning} into two subproblems: a \emph{learning}
subproblem, where a \emph{random geometric graph} $G$~\cite{penrose:rgg} meeting some
requirements is sought, and a \emph{combinatorial optimization} problem whose solutions are
paths $p \in Paths(G)$ such that a certain objective function, typically the sum of lengths of the
edges, is minimized~\cite{korte:combinatorial}.
Before discussing the PRM algorithm, which is given in Figure~\ref{alg:prm}, we formalize the learning subproblem as
that of searching over an infinite set of geometric graphs ${\cal G}$, looking for those
that meet the following criteria
\begin{subequations}
\label{learning_problem}
\begin{align}
\min_{G=(V,E) \in {\cal G}(S)} &\;\; D(V, S) \\ 
\suchthat &\;\; V\subset S,\, E \subset \Lambda \label{eq:rgg_struct}\\
&\;\; q_0 \in V, F_{\star} \cap V \neq \label{eq:rgg_cover} \emptyset
\end{align}
\end{subequations}
\noindent Solutions or \emph{roadmaps}, are random geometric graphs $G=(V,E)$ required, via constraint~\eqref{eq:rgg_struct}, to be such that the vertex set is a subset
of topological space $S$, and edges in $E$ correspond to functions $\lambda \in \Lambda$, $\lambda: [0,1] \to S$, or
\emph{local plans}. Solutions (graphs) $G$ are also required, via constraint~\eqref{eq:rgg_cover}, to cover the initial
state $q_0$ and the set of goal states $F_{\star}$. As noted by Janson et al.~\cite{janson:ijrr}, $\ell_2$-dispersion~\eqref{l2:dispersion}
for vertex set $V$ describes how well the former covers the set $S$, and therefore bounding how close any plan associated with 
a path $p \in Paths(G)$ can approximate the (unattainable) optimal cost for Problem~\eqref{generic_motion_planning}. 
Many formulations of~\eqref{learning_problem} exist in the
literature, with significantly different properties. 

\begin{remark}
In the original PRM paper~\cite{kavraki:94:roadmaps},
$S$ $=$ $F$ and $\Lambda$ $=$ $\{ \rho \in \Pi$ $:$ $\img{\rho} \subset F\}$, so that paths $p \in Paths(G)$ map directly onto solutions for Problem~\eqref{generic_motion_planning}.
In contrast, Bohlin and Kavraki set $S=C$ and $\Lambda$ to be the set of local plans $\lambda: [0,1] \to C$, so that paths $p \in Paths(G)$
may or not map onto solutions for Problem~\eqref{generic_motion_planning}.
\end{remark}

We adapt the classic descriptions of PRM (Figure~\ref{alg:prm}) in the literature to be an iterative algorithm, that generates
a sequence of geometric graphs $G_k$, $k=0,1,\ldots$. The initial graph $G_0$ is given by the tuple $(\{q_0, q_{\star}\}, \emptyset)$,
where $q_{\star}$ is a suitably chosen state in $F_{\star}$. Each iteration Algorithm~\ref{alg:prm} chooses $G_{k+1}$, a supergraph of $G_k$, by adding
a new vertex $q_k \in S$ and as many edges $\lambda \subset \Lambda$ as possible that connect $q_k$ with points $q' \in V_k$ in a suitably defined 
{neighborhood} centered at $q_k$. Line $2$ in Figure~\ref{alg:prm} initializes the iteration by \emph{sampling} state $q_k$ 
from a \emph{sequence} ${\cal X} \coloneqq X_0 X_1 \ldots X_k \ldots$ of \emph{random variables} whose distribution 
support\footnote{We note that the support of a random variable $X$ taking values in $\mathbb{R}^n$
is the set $\{ x \in \mathbb{R}^{n} \,:\, P_{X}({\cal B}_{r}(x)) >0,\,\mbox{for all}\, r > 0 \}$, and more generally, the smallest closed set such 
that $P_X(C)=1$, where $P_X(S)$ is the probability of $X$ taking as a value an element of set $S$.} is $S$. 
Line $3$ enumerates the elements in the neighborhood of $q_k$, which is defined by the set of existing vertices $V_k$ and
a ball of radius $r_k > 0$. This parameter, or \emph{connection radius}, decreases monotonically as $k$ increases, ensuring
that as $G$ becomes bigger, $\ell_2$-dispersion~\cite{janson:ijrr,solovey:20:prm} is reduced. Importantly,
the schedule for $r_k$ cannot be arbitrary as it greatly affects the \emph{convergence} of $G_k$ to a graph where $q_0$ and
some vertices $F_{\star} \cap V$ belong to the same connected component~\cite{solovey:20:prm}. Edges (local plans $\lambda$)
are added in lines $4$ and $5$ of Figure~\ref{alg:prm}. Depending on the definition of $\Lambda$ (see above), no valid local plan may exist, and 
then no edge is added.

\section{Revisiting Lazy PRM}
\label{revised_lazy_prm}

\begin{figure}[t]	
\centering
\includegraphics[width=0.4\textwidth]{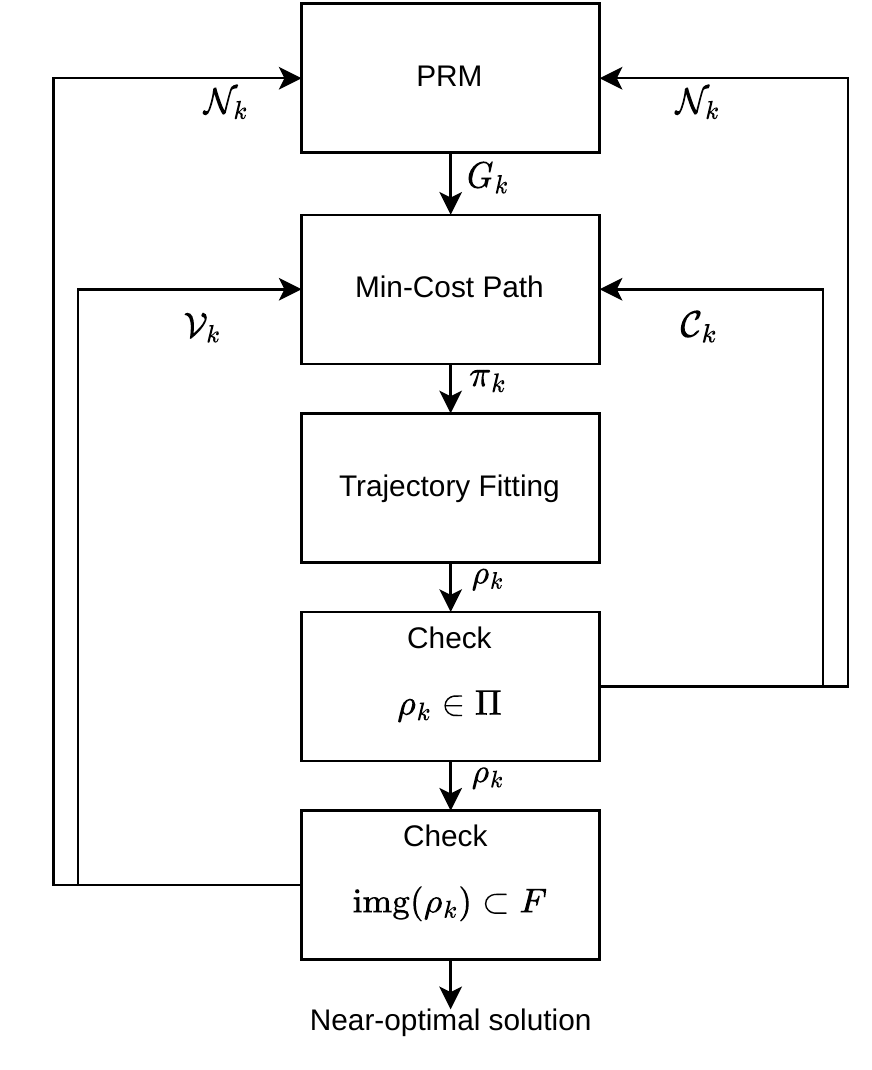}
\caption{Revised Lazy PRM, see text for details and discussion.}
\label{fig:revised_decomposition}
\end{figure}

We revisit the notion of Bohlin and Kavraki's Lazy PRM, proposing a different structuring into subproblems, 
as well as novel methodologies to address
them. These changes and additions are depicted in Figure~\ref{fig:revised_decomposition}. First, we
set $S = F$ and choose $\Lambda$ in Problem~\eqref{learning_problem} as the set of linear piecewise functions discussed in Example 1. From the resulting
geometric graph $G_k$, we formulate a Constrained Min-Cost Path (CMCP) problem (section~\ref{section:constrained_best_path}),
and seek its optimal solutions, paths $\pi_k \in Paths(G_k)$, which we will refer to as \emph{geometric plans}. 
These do not in general correspond with plans $\rho \in \Pi$, so 
we need to determine if $\pi_k$ can be mapped onto some element of $\Pi$. 
The existence of this mapping, specified on elements, is verified in two steps. First, we choose a smooth function $\rho_k$, which
we will be referring to as \emph{smooth plan}, by solving
a (convex optimization) \emph{trajectory fitting} problem~\cite{boyd:04:convex} (section~\ref{section:trajectory_fitting}).
This choice is constrained so that $\rho_k$ is $C^2$-continuous and traces through each vertex
in $\pi_k$. Second, we then check if suitably defined kinematic constraints on $\rho_k$ are satisfied,
certifying that there is a match between $\pi_k$ and at least one element in $\Pi$ 
(section~\ref{velocity_and_kinematic_constraints}).
If this check fails, and $\rho_k$ is proven to not be an element of $\Pi$, a cut ${\cal C}_k$ is generated and added to the set of 
constraints of the CMCP problem. Proving that $\rho_k \in \Pi$ is not sufficient, as we need to check if $\img{\rho_k}\subset F$ 
(section~\ref{efficient_and_exact}), ensuring that both kinematic and geometric constraints are satisfied simultaneously. If this 
second test fails, a new cut ${\cal V}_k$ is obtained and added to the constraints of the CMCP. Both of these tests implement the
cutting plane component of our BC algorithm. Finally, whenever the check $\rho_k \in \Pi$ (resp. $\img{\rho} \subset F$) is
passed by the candidate smooth plan $\rho_k$, cuts ${\cal C}_k = \emptyset$ (resp. ${\cal V}_k = \emptyset$).

\begin{remark}
	In this paper it is assumed that checking local plans (straight lines) $\lambda \subset F$ to obtain the geometric graphs $G_k$ 
	in Figure~\ref{fig:revised_decomposition} is computationally efficient. When this assumption does not
	hold, one can revert to the formulation of Problem~\eqref{learning_problem} used by Bohlin and Kavraki.
	The algorithm in Figure~\ref{fig:revised_decomposition} does not need to change, as the procedure described in 
	section \ref{efficient_and_exact} is guaranteed to reject plans $\rho_k$ s.t. $\img{\rho} \cap O \neq \emptyset$.
\end{remark}

The algorithm in Figure~\ref{fig:revised_decomposition} iterates for a set amount of time, possibly finding several 
valid plans $\rho_k$ with monotonically decreasing costs. The selection of geometric graphs $G_k$ is steered
by a sampling sequence ${\cal X}$ of random variables $X_k$. Each of these chooses, with equal probability, either points $q \in F$
according to distribution $P_{X_k}(F) = 1$, or points $q \in {\cal N}_k \subset F$ from \emph{search direction} ${\cal N}_k$. 
These are defined in terms of the vertices featuring in cuts ${\cal V}_k$ and ${\cal C}_k$ generated so far
(section~\ref{no_goods_and_branching}). This heuristic,
analogous to the notion of \emph{activity-based} branching heuristics~\cite{moskewicz:chaff} in discrete optimization, ensures that geometric graphs $G_k$ with paths
that are structurally similar to those rejected by previous cuts are frequently considered without compromising completeness.

\section{Constrained Min-Cost Path Problems}
\label{section:constrained_best_path}

We start by introducing the formulation of minimum-cost path (MCP) as an Integer Linear Program (ILP)~\cite{wolsey:ip} 
where costs are defined over the edges in a path. Given a
geometric graph $G=(V,E)$, the task is to find the path 
$p=q_0 q_1 \ldots q_i \ldots q_N q_t$ in the graph $G' = (V', E')$ that follows from adding a dummy vertex $q_t$ and additional
edges $\{ (q_i, q_t) \,:\, q_i \in V \cap F_{\star}\}$, and
that minimizes $f(p)$, a linear function. 
The set of 
all possible paths is represented by a vector of \emph{decision variables} 
$x \in \{0, 1\}^{\vert V \vert^2}$, so each $x_{ij}$ allows to choose
whether edge $(q_i, q_j)$ is in the solution path $p$. The objective $f(p)$ is defined as the inner product
of $x$ with a vector $c \in \mathbb{R}^{\vert V \vert^2}$ of \emph{edge costs}. These
costs can be, for instance, the \emph{arc-length} of the local plan $\lambda_{ij}$ connecting $q_i$ and $q_j$.
The choice of which edges to use is restricted by a number of constraints that follow from the structure
of the graph $G$, so the \emph{optimal cost} $\phi(G)$ of the MCP for graph $G$ instance is described by the following mathematical program
\begin{subequations}
\label{min_cost_path}
\begin{alignat}{2}
\phi(G) \coloneqq \min_{x} &\; \sum_{i=1}^{\vert V \vert} \sum_{j=1}^{\vert V \vert} c_{ij} x_{ij} & \label{sp:objective} \\
\suchthat &\; \sum_{(q_0,q_j) \in E} x_{0j} = 1 & \label{sp:c1} \\
& \; \sum_{(q_i,q_k)\in E} x_{ik} - \sum_{(q_k,q_j)\in E} x_{kj} =0 \label{sp:c2}\\
& \;\sum_{q_i \in V_{\star}} x_{it} = 1 & \label{sp:c3} \\
& \; x_{ij} = 0,\, \mbox{for}\,i,j,\mbox{s.t.}\,(q_i,q_j) \notin E \label{sp:c4}
\end{alignat}
\end{subequations} 
\begin{remark}
The above is a trivial variation on the classic formulation of shortest-path problems in graphs
as a special case of minimum-cost flow problems~\cite{kipp-martin:linear}.
\end{remark}
The objective function~\eqref{sp:objective}
is a linear function as introduced above, where each coefficient $c_{ij}$ corresponds to the value of
a cost function $f: \Lambda \to \mathbb{R}$. Constraint~\eqref{sp:c1} follows from the trivial fact that all valid paths
must start at vertex $q_0$, therefore exactly one of $q_0$ outgoing edges must be in the path.
\emph{Flow constraints}~\eqref{sp:c2} range over every $q_k \in V$ but $q_0$,
and ensure that if an incoming edge of $q_k$ is in the selected path $p$, then one of its outgoing
edges must also be in $p$.  Constraint~\eqref{sp:c3} ensures that the path $p$ ends at the dummy vertex $q_t$,
and constraint~\eqref{sp:c4} prevents selecting into $p$ possible edges for which no local plan has been 
found.

\begin{remark}
ILPs like Problem~\eqref{min_cost_path} can be efficiently solved via generic algorithms for Integer Linear and Dynamic Programming 
based on Branch-and-Bound~\cite{kipp-martin:linear,bertsekas:17:dp}. The choice of algorithm being determined by (1) the nature of
the coefficients of the cost function, (2) availability of good cost relaxations to orient the search for solutions, (3) the 
maximal number $b$ of outgoing edges from any vertex in $V$, or
\emph{branching factor}, and (4) the least number $d$ of actions (edges) in optimal plans, or \emph{depth} of optimal solutions.
\end{remark}

\subsection{Branch-and-Cut over Probabilistic Roadmaps} 
\label{branch_and_cut}
As noted in section~\ref{revised_lazy_prm}, the definition of $S$ and $\Lambda$ in Problem~\eqref{learning_problem}, determines whether
there exists an injective mapping between solutions $\pi$ to constraints~\eqref{sp:c1}--\eqref{sp:c4} and plans $\rho$ that solve 
Problem~\eqref{generic_motion_planning}. Recovering this property would require us to augment Problem~\eqref{min_cost_path} with constraints
\begin{align}
\label{no_goods}
\sum_{(q_i, q_j) \in {\cal E}}x_{ij} \leq \vert {\cal E} \vert - 1
\end{align}
for each possible subset ${\cal E} \subset E_k$ of the edges in graph $G_k$ that (1) appear contiguously in some solution
of Problem~\eqref{min_cost_path} and (2) there is no $\rho \in \Pi$ tracing the vertices in ${\cal E}$. Doing so
is only feasible for very small graphs, as the number of sets ${\cal E}$ to consider is exponential on the number of 
vertices of $G_k$. Besides the sheer number of edge sets to test and constraints to be added, only a very small fraction of these
will be relevant to optimal solutions of Problem~\eqref{min_cost_path}, so the computational effort invested in verifying every 
path and sub-path is not amortised. Avoiding
such wasted effort is one of the key motivations for us to look again into Lazy PRM.

Branch-and-cut (BC) was originally proposed in the context of Mixed-Integer Linear Programming~\cite{mitchell:branch_and_cut} as a 
way to manage combinatorial explosions in the number of sub-tour elimination constraints~\cite{grotschel:subtour} required to 
formulate the Travelling Salesman Problem (TSP). We observe that BC is directly applicable to motion planning over probabilistic roadmaps,
generalizing the results of Bohlin and Kavraki, as the problem of selecting best paths in $G_k$ can be trivially described by an integer 
linear problem. Concrete BC algorithms for
motion planning thus need to (1) decide what constraints need to be \emph{separated} from the ones used to identify 
optimal solutions, and (2) specify one or more procedures to check when an optimal solution for Problem~\eqref{min_cost_path}
violates any of the separated constraints, generating \emph{cuts}~\eqref{no_goods} that are added to~\eqref{sp:c1}--\eqref{sp:c2}. 
We discuss these procedures and the cuts generated in section~\ref{section:constraint_checking}.

\section{Trajectory Fitting}
\label{section:trajectory_fitting}

The next subproblem we address is that of \emph{trajectory fitting}, in which we map geometric, $C^0$-continuous plans $\pi_k$, 
a sequence of local plans $\lambda$ selected by optimal solution of CMCP problems, into smooth, $C^2$-continuous plans $\rho_k$.
We approach this problem as a \emph{function fitting} problem~\cite{boyd:04:convex}, where we select a member of a finite dimensional
subspace of functions, a \emph{superset} of plan domain $\Pi$ in Problem~\eqref{generic_motion_planning}, that is the \emph{best fit} given
some data (waypoints $q_i$, $i=0,\ldots,N$ in $\pi_k$) and requirements (continuity of the derivatives at each waypoint). Formally, given a 
family of \emph{basis functions} $\eta_1, \ldots, \eta_n : \mathbb{R} \to \mathbb{R},$ we seek a function
\begin{align}
\label{best_fit}
\eta(t) = \theta_{1} \eta_{1}(t) + \theta_{2} \eta_2(t) + \ldots + \theta_{n} \eta_n(t),
\end{align}
where the coefficients $\theta_1,...,\theta_n \in \mathbb{R}^d$ parameterize the subspace of functions and become the decision variable for an
optimization problem. In line with the mathematical tractability and robot controllability concerns expressed in the Introduction, 
we consider polynomial basis functions
\begin{align}
\label{polynomial_basis}
\eta_{l}(t) = t^{l-1},\,l=1,\ldots,n
\end{align}
matching the requirements stated in section~\ref{motion_planning}. We now observe that the geometric plan $\pi_k$ is defining a
\emph{triangulation} of the domain of plans (functions) $\rho$, the interval $[0,1]$. That is, geometric plan $\pi = q_0 q_1 \ldots q_i \ldots q_N$
is partitioning $[0,1]$ into $N$ disjoint \emph{simplices} (lines) comprised between the grid points 
$g_0 = 0, g_1 = 1 / N, \ldots, g_i = i/N, \ldots, g_N=1$. A piece-wise polynomial then can be obtained directly by fitting to each
of the simplices a function like~\eqref{best_fit} that is defined over the basis~\eqref{polynomial_basis}, requiring the desired
continuity constraints at the boundaries between each simplex. In this paper we consider, the 
polynomial basis with $l=4$, that is, smooth plans $\rho$ are \emph{cubic splines}. The constraints and analysis that follows can be
trivially extended into higher degree basis (like $n=6$), or other types of $C^2$-continuous spline curves, like B-splines~\cite{sprunk:splines}. 

Let $\theta = (\alpha, \beta, \kappa, \delta) \in \mathbb{R}^4$, then for each dimension $j=1,\ldots,d$ the polynomial piece $a_i$, $i=0,\ldots,N-1$,
is defined in terms of parameters $\theta_{ij}$ as follows:
\begin{align}
\label{cubic_spline}
a_{ij}(t) \coloneqq \alpha_{ij}t^3 + \beta_{ij}t^2 + \kappa_{ij}t + \delta_{ij},\, j = 1, \ldots, d
\end{align}
For each simplex $i=0,\ldots,N-1$ and $j=1,\ldots,d$ we require that the endpoints match adjacent waypoints, e.g. $\rho_{ij}(0) = q_{ij}$ and $\rho_{ij}(1) = q_{i+1j}$,
which translates into the following linear constraints
\begin{align}
\delta_{ij} - q_{ij} = 0 \label{fit:c1} \\
\alpha_{ij} + \beta_{ij} + \kappa_{ij} + \delta_{ij} - q_{{i+1}j} = 0 \label{fit:c2}
\end{align}
$C^2$-continuity is enforced by requiring that $\dot\rho_{ij}(1) = \dot\rho_{i+1j}(0)$, and
$\ddot\rho_{ij}(1) = \ddot\rho_{i+1j}(0)$ for $i=0,\ldots,N-2$ and $j=1,\ldots,d$, that translates into constraints
\begin{align}
3 \alpha_{ij} + 2 \beta_{ij} + \kappa_{ij} - \kappa_{i+1j} = 0 \label{fit:c3} \\
6 \alpha_{ij} + 2 \beta_{ij} - 2 \beta_{i+1j} = 0 \label{fit:c4}
\end{align}
Our last requirement is specific to the application of function fitting to robot motion planning, and
it requires acceleration to be 0 at the initial and final state visited by the plan $\rho$. This
requirement constraints are
\begin{align}
2 \beta_{0j} = 0 \label{fit:c7} \\
6 \alpha_{N-1j} + 2 \beta_{N-1j} = 0 \label{fit:c8}
\end{align}
It is easy to see that constraints~\eqref{fit:c1}--\eqref{fit:c8} form a system of linear equations with
$4d(N-1)$ variables and constraints. When the constraints are linearly independent and consistent, there 
is a unique solution $\rho = a_0$, \ldots, $a_i$, \ldots, $a_{N-1}$.

\section{Constraint Checking and Search Direction}
\label{section:constraint_checking}
We now define the checking procedures to produce cuts ${\cal C}_k$ and ${\cal V}_k$, and how
these are used to define the search directions used to select geometric graphs.

\subsection{Exact Constraint Checking for Smooth Plans}
\label{efficient_and_exact}

The \textsc{PolyTrace} algorithm \cite[Algorithm 1]{selvaratnam:polynomial_verification} provides the 
means to test whether $\mathcal{B}_{\delta_{i}}(\rho) \cap O_i = \emptyset$, for a given polynomial 
$\rho:[0,1] \to \mathbb{R}^d$ and region $O_i \subseteq \mathbb{R}^d$ described by \eqref{eq:Oi}. 
The test assumes that clearances $\delta_i$ are rational, and that all polynomials involved have rational 
coefficients: that is, $\rho \in \mathbb{Q}[t]^d$, each $g^i_j \in \mathbb{Q}[x_1,...,x_d]$, 
and $\delta_i \in \mathbb{Q}$.

Let us start by ignoring clearance constraints. The polynomial $\rho$ intersects $O_i$, 
i.e. $\mathrm{img}(\rho) \cap O_i \neq \emptyset$, iff exists $k$,  
\begin{align}
\zeta(k) = \{1,...,m_i\} \label{eq:checkTrace}
\end{align}
where $\zeta := \textsc{PolyTrace}(\rho,(g^i_1,...,g^i_{m_i}))$ is 
a \emph{trace}~\cite[Definition 19]{selvaratnam:polynomial_verification}, thus 
recording every region visited by $\rho$. To visit $O_i$, the polynomial $\rho$ must 
simultaneously visit $\{ x \in \mathbb{R}^d \mid g^i_j(x) \leq 0 \}$ for every $1 \leq j \leq m_i$, 
as indicated by \eqref{eq:checkTrace}. 
We now reintroduce clearance requirements. We observe that
\begin{align}
\mathcal{B}_{\delta_i}(\varphi) \cap O_i = \emptyset \iff \img{\rho} \cap \left( O_i \oplus \mathcal{B}_{\delta_i}(0) \right) = \emptyset
\end{align}
That is, the clearance constraint is satisfied iff $\rho$ does not intersect $O_i \oplus \mathcal{B}_\delta(0)$, a $\delta_i$-\emph{expansion} 
of $O_i$. If we can find polynomials $h^i_j \in \mathbb{Q}[x_1,...,x_d]$ such that 
\begin{align} 
O_i \oplus \mathcal{B}_\delta(0) \subseteq \bigcap_{i=1}^{m_i} \{ x \in \mathbb{R}^d \mid h^i_j(x) \leq 0\}, \label{eq:expansionContainment} 
\end{align}
then the problem of determining whether $\mathcal{B}_{\delta_{i}}(\rho) \cap O_i$ is solved by examining $\zeta := \textsc{PolyTrace}(\varphi,(h^i_1,...,h^i_{m_i}))$. If \eqref{eq:checkTrace} fails to hold, then $\mathcal{B}_{\delta_i}(\rho) \cap O_i = \emptyset$, and sufficient clearance is guaranteed. 
In general, finding such $h^i_j$ can be challenging
But if the $g^i_j$ are affine, then \eqref{eq:expansionContainment} holds when we set
\begin{align}
h^i_j(x):= g^i_j(x) - \delta_i c \label{eq:expanded_constraints}
\end{align} 
for any $c \geq \| \nabla g^i_j(x) \|$. Recall that $\| \nabla g^i_j(x) \|$ is constant because $g^i_j$ is affine. If 
we define $c := \| \nabla g^i_j(x) \|$, then \eqref{eq:expansionContainment} holds with equality, but then $c$ is typically irrational, and leaves $h^i_j$ having 
irrational coefficients. Choosing $c \in \mathbb{Q}$ to be a rational upper bound on $\|\nabla g^i_j(x) \|$ instead, 
guarantees $h^i_j \in \mathbb{Q}[x_1,...,x_d]$, as required. This introduces some conservativeness into the test. 

We generate cuts ${\cal V}_k$~\eqref{no_goods} when we find an action $a_j$ in $\rho$ that satisfies condition~\eqref{eq:checkTrace}, and ${\cal V}_k$ is
defined as
\begin{align}
\label{eq:validity_cut}
{\cal V}_k \coloneqq \{ e_l \in \pi_k\,:\, j_{-} \leq l \leq j_{+} \}
\end{align}
where $j_{-} = \max\{ 1, j-1 \}$, $j_{+} = \min\{ j+1, N\}$ and $N = \vert \pi_k \vert$. If no such action $a_j$ exists, then the statement 
$\img{\rho} \subset F$ is proven to be true.
\begin{remark}
Requiring every $g^i_j$ to be affine restricts the regions $O_i$ to either be polytopes, or the result
of over-approximating each $O_i$ with a polytope. 
\end{remark}

\subsection{Managing Arithmetic Complexity}
\label{managing_arithmetic}
Applying \textsc{Polytrace} in the manner described in the previous section requires careful consideration
of the arithmetic complexity of the calculations required. We indentify and address two main challenges.
First, all coefficients in constraints $O_i$ and plans $\rho$ must be rationals. Secondly, the time
and space complexity  of \textsc{Polytrace}~\cite[Algorithm 1]{selvaratnam:polynomial_verification}
is dominated by that of finding the isolating intervals~\cite{melhorn:vca} for the roots of $N$ products
of polynomials that result from the composition of the right-hand side of constraints and each action $a_j \in \rho$. 
The result of this arithmetic operation, yet another polynomial, sometimes have very high degree and
very large coefficients that cannot be represented with fixed-size data types native to computer programming
languages. 
The parameters governing the growth of degree and maximal coefficient size of this product depend on the degree of the
basis (section~\ref{section:trajectory_fitting}) used to define $\rho$ and $m_i$. For instance, when using polytopes
to represent constraints $O_i$ and using the basis~\eqref{polynomial_basis} with $l=4$ on the instances discussed in
section~\ref{illustration_and_experiments}, we obtain fairly 
frequently polynomials of degree up to $16$ with coefficients consisting of dozens of digits.
 
The first problem is overcome by computing rational approximations of coefficients of constraints $O_i$ and
actions $a_j$ in plans $\rho_k$. Each floating-point number $\xi$ representing a coefficient is mapped into
a \emph{dyadic rational}~\cite{knuth:acp2} $\xi'$, where the denominator is $d \in \mathbb{N}$, $1 \leq d \leq 2^\beta$
\footnote{In this paper, $\beta=14$, which results in an approximation down to the fourth decimal place of floating-point numbers $\xi$.}.
$\xi'$ is selected so as to minimize $\vert \xi - \xi' \vert$ and the bitsize of $d$.
Using small powers of two
as denominators greatly reduces the bit complexity of the arbitrary precision arithmetic operations required to implement 
\textsc{Polytrace}\footnote{We use the \texttt{boost::cpp\_int} library for this.}.

\begin{remark}
	Alternatively, and when performance is not a concern, a 
	Computer Algebra System such as \textsc{Mathematica}~\cite{mathematica} or \textsc{SymPy}~\cite{sympy} can be readily used to implement \textsc{Polytrace}.
\end{remark}

The second problem required to analyze how the truth of each \emph{relational atom} $\alpha_{il} \equiv \big( g_{i}^{l} \circ \rho_j(t) \big) \leq 0$ changes
through the actions $\rho_j$ in the incumbent smooth plan. For that we borrow the notion of \emph{inertial atoms}~\cite{giunchiglia:causal_logic},
and observe that existence of \emph{any} roots for a polynomial $\rho_j$ in a interval $I$ can be determined by means of 
\emph{Descartes' sign rule}~\cite{melhorn:vca}. Thus, for every action $\rho_j$ in the plan, we can evaluate the truth of $\alpha_{il}$ at $a_j(0)$ and $a_j(1)$,
and if Descartes' rule certifies that no roots exist in the interval $(0,1)$, then we know $\alpha_{il}$ is \emph{inertial}, e.g.
its truth value remains constant within the $[0,1]$ interval, and does not need to be considered in any calls to \textsc{Polytrace} to obtain 
traces $\zeta_j$. This simple and efficient test very often greatly reduces the complexity of product polynomials when $O_i$ is a polytope.

\subsection{Velocity and Kinematic Constraints}
\label{velocity_and_kinematic_constraints}

Following~\cite[Chapter 13]{lavalle:06}, we now assume that $C$ specifies the possible configurations
of one or more rigid bodies. As discussed in section~\ref{motion_planning}, plan domain $\Pi$ is given
by $m'$ \emph{implicit differential constraints}~\ref{plan_domain_def} on plans $\rho: [0,1] \to C$.
As long as each $g_i \circ \dot\rho$ in~\ref{plan_domain_def} are polynomials in $\mathbb{R}[t]^d$, we can use \textsc{Polytrace}
to generate a trace $\zeta$ as we did to check geometric constraints in section~\ref{efficient_and_exact}.
We then can establish that $\rho \in \Pi$ if for all $k=1,\ldots,\vert \zeta \vert$ it holds that
$\vert \zeta(k) \vert = m$. In words, at no point $t \in [0, 1]$ we found $\rho(t)$ to be somewhere else than
inside the region in tangent space enclosed by the boundaries of constraints $g_i$ in~\eqref{plan_domain_def}.
A cut constraint ${\cal C}_k$~\eqref{no_goods} follow from actions $a_j$ in $\rho$ that do not satisfy~\eqref{plan_domain_def}
\begin{align}
\label{eq:kinematic_cut}
{\cal C}_k \coloneqq \{ e_l \in \pi_k \,:\, j_{-} \leq l \leq j_{+} \}
\end{align}
where $j_{-}$, $j_{+}$ and $N$ are as in~\eqref{eq:validity_cut}.

When $g \circ \dot\rho$ is not a polynomial, we can still define useful cuts, but only at specific
points along plans $\rho$. We illustrate this with the following constraints. Let us suppose $C=\mathbb{R}^2$,
and $\rho$ is a piece-wise $C^2$-continuous function obtained as per Section~\ref{section:trajectory_fitting}.
We set the following constraint for each action $a_j$ in $\rho$, and intervals $I=(l,r)$ $\in$ $\{ (0, 1/2)$, $(1/2, 1) \}$
\begin{align}
\label{eq:curvature_constraint}
-\phi_{max} \leq \mathrm{arccos}\bigg(\frac{\dot a_j(r) \cdot \dot a_{j}(l)}{ \Vert \dot a_{j}(r) \Vert \Vert \dot a_{j}(l) \Vert}\bigg) \leq \phi_{max}
\end{align} 
where $\phi_{max}$ is the largest allowed 
change in the direction of the tangent of plan $\rho$ along actions
$a_j$. Constraint~\eqref{eq:curvature_constraint} is useful to model limits on steering angles for a wide family
of robotic systems~\cite{laumond:98:guidelines}. If some action(s) $a_j$ do not satisfy constraints~\eqref{eq:curvature_constraint}
then cuts ${\cal C}_k$ as above~\eqref{eq:kinematic_cut} are generated.

\subsection{Directing the Search for Geometric Graphs}
\label{no_goods_and_branching}

As advanced in section~\ref{revised_lazy_prm} we use a dynamic sampling sequence ${\cal X}$ 
(section~\ref{probabilistic_roadmaps}) where random variables $X_k$ for iteration $k$ are 
use the information conveyed by cuts ${\cal V}_j$ and ${\cal C}_j$, for $j < k$,
directing the selection of graph $G_k$.

For each cut ${\cal V}_j$ and ${\cal C}_j$ generated up to iteration $k$, \emph{search directions} (distributions)
${\cal N}_j$ are defined following Bohlin and Kavraki~\cite{bohlin:lazy_prm:00},
but with \emph{seed states} $q_j$ defined to be the barycenter of vertices $q_{i}, q_{i'} \in V_j$
at endpoints of edges $x_{ii'}$ in cuts~\eqref{no_goods}. This heuristic exploits the
observation that cubic splines lack \emph{localism}~\cite{sprunk:splines}. By choosing
new vertices from ${\cal N}_j$ we will be generating plans $\rho_k$ that are close perturbations
of previously rejected plans $\rho_j$. 

\section{Evaluation}
\label{illustration_and_experiments}

\subsection{Benchmarks and Execution Environment}
\label{benchmarks}

We illustrate and evaluate the performance of our Lazy PRM algorithm on the BARN dataset, a publicly
available benchmark for robotic motion planning tasks~\cite{perille:20:benchmarking}
that also have a clear pathway for implementation on easily accessible high-fidelity simulation and robotic hardware. We use the $300$
instances available here~\cite{barn:website}, that represent freespace $F$ using Robot Operating System (ROS)~\cite{ros} 
occupancy grids. We map these into sets of (possibly overlapping) polytopes 
automatically via a simple greedy approximation of the problem given in \cite[Section 2]{bemporad:polytope_approx}.
This procedure is written in \texttt{Python}, and we observed it takes well below one second to process each instance
in the set~\cite{barn:website}. The size of the resulting geometric constraint sets $O$ ranges from $30$ to $70$ 
sets $O_i$, their boundaries described by $4$ to $6$ affine constraints. We impose constraint~\eqref{eq:curvature_constraint}, with maximum angle $\phi_{max}$ is $\pi/2$ (90$^\circ$). Initial and goal states are the same in every instance and chosen so 
that non-trivial tasks (e.g. $\rho$ has $N>1$ actions) are generated with high probability\footnote{We refer the reader to 
the videos available here: \url{https://bit.ly/3rbGPGr}}. 

We ran our planners (written in \texttt{C++}) in Ubuntu 20.04, over an 8-core 11th Generation Intel Processor with a clock speed of 3.6GHz. For each
instance (e.g. definition of $F$, $q_0$ and $F_{\star}$), we do $5$ runs, using numbers  $1$, $42$, $567$, $1337$ and $8193$ to seed random number generators. Runs are terminated 
after $500$ seconds or earlier if a suboptimal solution of Problem~\eqref{generic_motion_planning} is found. 
Reported timing information follows from measurement by the 
\texttt{steady\_clock} available through the \texttt{chrono} C++ standard library module. Runs working set is limited to $4$ GBytes
of RAM.

\subsection{Planners}
\label{planners}

We have tested two implementations of the Lazy PRM scheme described in Figure~\ref{fig:revised_decomposition}, that differ in the
choice of algorithm to solve the CMCP problem described in section~\ref{section:constrained_best_path}. The first algorithm is
Google's \textsc{Cp-Sat} Branch-and-Bound solver~\cite{ortools} that combines state-of-the-art techniques for Constraint Programming (CP) 
and Mixed-Integer Programming (MIP), using Lazy Clause Generation (LCG)~\cite{ohrimenko:lcg}. The second algorithm is
a bespoke two-tiered Dynamic Programming (DP) algorithm that uses Euclidean distance as a lower bound~\cite{pearl:heuristics}. This algorithm 
uses $A^{*}$~\cite{hart:68:astar} to solve instances of Problem~\eqref{min_cost_path}, and switches to Best-First Search~\cite[Chapter 5]{bertsekas:17:dp} 
when cuts~\eqref{no_goods} have been added. The latter algorithm lazily checks that cuts are satisfied before a path (node) selected from the 
search frontier is expanded.

\begin{remark}
	\label{a_star_remark}
	The reason for switching from $A^{*}$ follows from the observation that pruning paths $p \in Paths(G)$ from vertex $q_0$ when these do
	not satisfy some active cut~\eqref{no_goods} violates a key assumption in $A^{*}$. To wit, whenever a partial path $p$ between $q_0$ and
	a vertex $q$ is added to the open list, we will not need to consider any other path from $q_0$ that visits $q$, and thus any such
	alternatives can be pruned away. This is violated when ${\cal V}_k$ contains multiple edges, as the vertex $q$ may be reached through
	different edges, not featured on any active cut constraint. In order to use $A^{*}$, one would need to consider an extended graph $G' = (V', E')$
	defined as follows. The vertex set $V'$ would be given by $V' \coloneqq V \times \mathbb{Z}_{0}^{k+}$, the Cartesian product between
	the vertex set of the geometric graph and the set of integer vectors with non-negative elements, and initial vertex $v_0 \in V'$ 	
	is the pair $(q_0, \mathbf{0})$. Given cuts ${\cal V}_k$ and ${\cal C}_k$, 
	the edge set $E' \subset V' \times V'$	would consist of edges $e=(v, v')$ where $v=(q,z)$ and $v'=(q',z')$ that satisfy the
	following conditions:
	\begin{enumerate}
		\item vertices $q$ and $q'$ were connected already in $G$, e.g. $(q, q') \in E$,
		\item $z_{j}' = z_{j} + 1$ if $(q,q') \in {\cal V}_j$, or $(q, q') \in {\cal C}_j$, for $1 \leq j \leq k$, otherwise $z_j = z_{j}'$,
		\item $z_{j}' \leq \max\{ \vert {\cal V}_j \vert, \vert {\cal C}_j \vert \}$ for $1 \leq j \leq k$.
	\end{enumerate}
	Constructing these augmented graphs $G'$ imposes significant computational burdens, even when using compact representations for
	$E$ such as STRIPS operators~\cite{fikes:71:strips,frances:15:icaps} as the geometric graphs $G$ and number of cuts $k$ to consider grows 
	with each iteration. Also, sophisticated data structures become necessary to implement efficiently the open and closed lists.
\end{remark}

For each planning algorithm we tested two different definitions of ${\cal X}$. In addition to any generated neighborhoods
${\cal N}_j$, $j=0,\ldots,k-1$, we consider two undirected sampling strategies. One is to sample a uniform distribution over $F$, 
the other is to enumerate the Halton sequence. We compare Bohlin and Kavraki~\cite{bohlin:lazy_prm:00} definition of
seed states, with the one proposed in section~\ref{no_goods_and_branching}, as the latter considers multiple edges in generated cuts.
We start by generating a geometric graph $G_0$ with $20$ vertices, and whenever no solutions are found for 
Problem~\eqref{min_cost_path} over graph $G_k$, 
we add $10$ new vertices to obtain $G_{k+1}$, $5$ from a randomly chosen neighborhoods ${\cal N}_j$, the other from the undirected
samplers.

\subsection{Overview of Results}
\label{overview}

\begin{figure}[ht]
\centering
\includegraphics[width=0.6\textwidth]{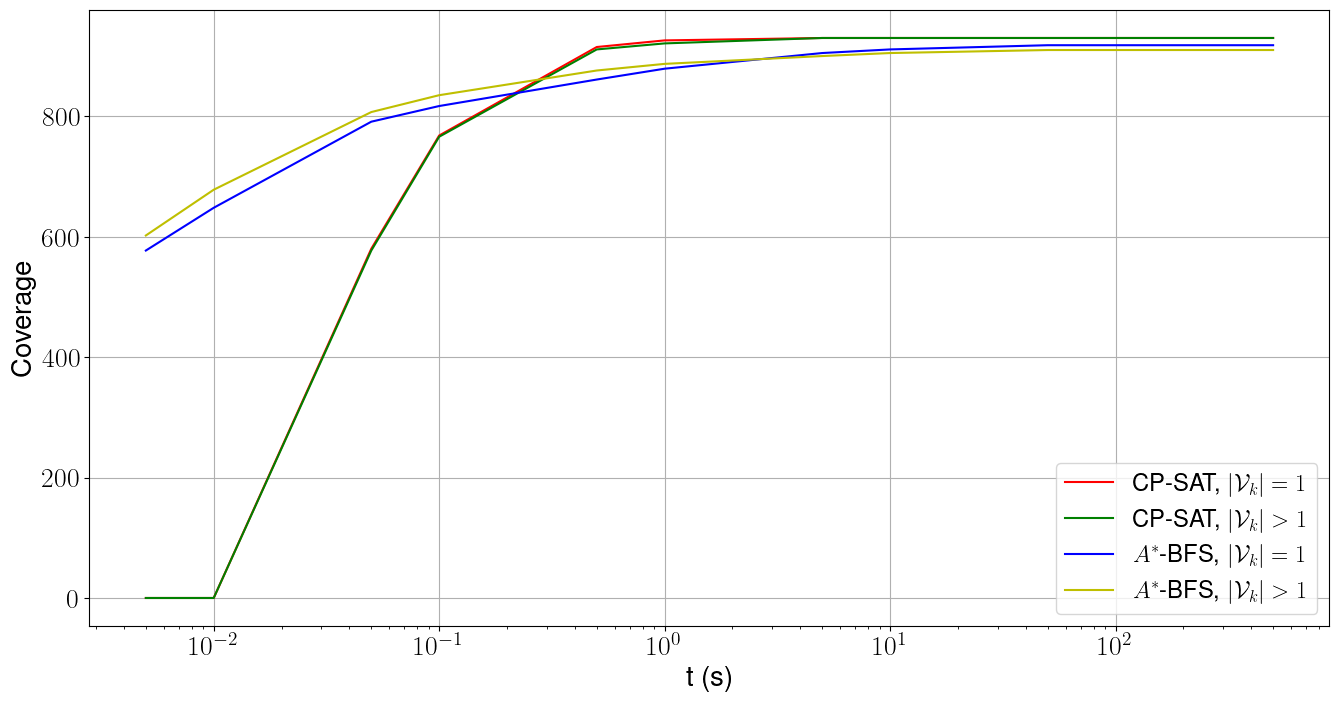}	
\caption{Coverage (number of runs finding one solution) over time for CP-SAT and $A^*$-BFS planners, using deterministic undirected sampling, and either
single or multiple-edge cuts. See text for discussion. }
\label{fig:coverage_comparison}
\end{figure}
Figure~\ref{fig:coverage_comparison} show the \emph{coverage over time} curves for the planners with highest coverage (number of solved instances
within time limits) and best quality plans. Planners based on CP-SAT solved all $1,500$ instances, and we found DP-based ones to run out of memory
in about $1\%$ of them. This event is significantly more frequent when using uniform undirected sampling. CP-SAT has a significant overhead when
starting up: this follows from us relying on
the (default) solver interface designed for distributed applications, 
that marshalls the variables and constraints of Problem~\eqref{min_cost_path} to be transmitted over a network. Much lower initial latency
is possible by using directly CP-SAT low-level data structures. Trajectory fitting (section~\ref{section:trajectory_fitting}) and checking of kinematic
constraints~\eqref{eq:curvature_constraint} have negligible impact on runtimes. Geometric constraint verification takes up 80\% of the runtime
on average, the other 20\% being spent finding minimum-cost paths over $G_k$.
\begin{figure}[ht]
\centering
\includegraphics[width=0.6\textwidth]{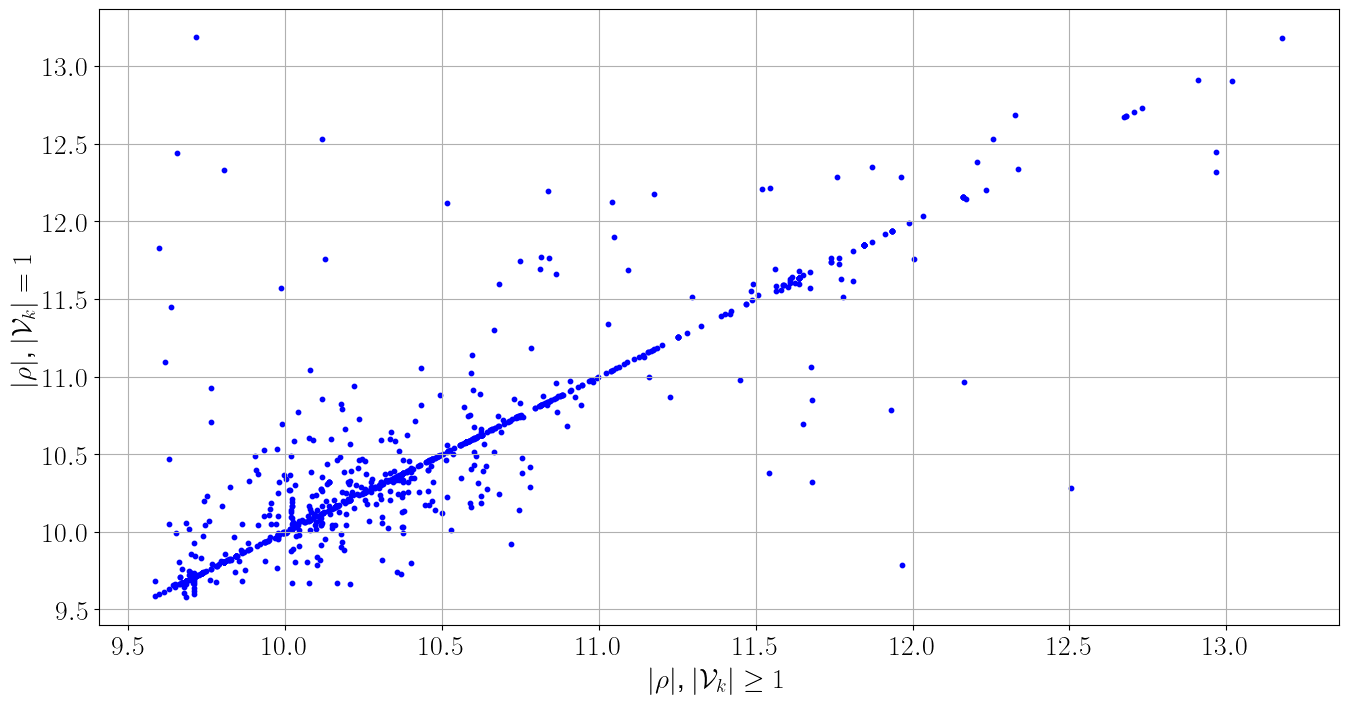}	
\caption{Approximate arc-lengths of plans $\rho$ for multi (x-axis) and single-edge (y-axis) cuts ${\cal V}_k$,
when using deterministic undirected sampling.}
\label{fig:cost_comparison}
\end{figure}
We observed a significant difference when using single and multiple-edge cuts ${\cal V}_k$ when it comes to plan costs, obtained from the numeric integration
of the arc-length for smooth plans $\rho$. Figure~\ref{fig:cost_comparison} shows that for most of runs, there is no cost difference, but
when using larger cuts ${\cal V}_k$ lower cost plans are often obtained. More often we observe that using larger cuts results in smaller differences
in costs w.r.t. the minimum cost observed. In Figure~\ref{fig:cost_comparison} we show the results when using the Halton sequence to define ${\cal X}$,
and the observed trend is noticeably amplified when using uniform random sampling.
\begin{figure}[ht]
\centering
\includegraphics[width=0.6\textwidth]{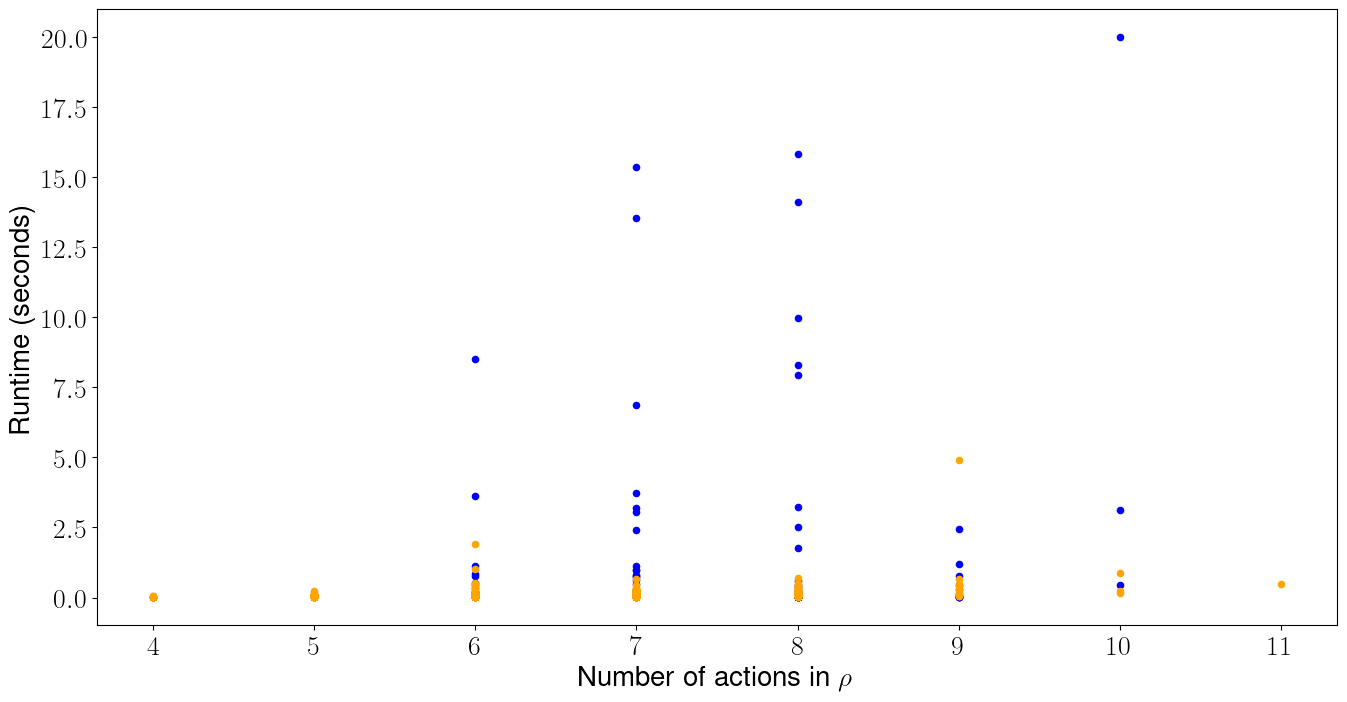}	
\caption{Number of actions in plans $\rho$ and run-time of planners. Blue is for $A^{*}$-BFS planner, orange for CP-SAT.}
\label{fig:plan_len_comparison}
\end{figure}
Finally, we compare in Figure~\ref{fig:plan_len_comparison} $A^*$-BFS and CP-SAT runtimes as the number of actions in plans $\rho$ increases. 
DP planners are significantly faster in runs where a valid plan $\rho$ was found after a few iterations, graphs are small and plans do not have many actions 
(75\% of instances admit valid plans with $\leq 6$ actions, and the shortest $\rho$ observed has $4$ pieces).
CP-SAT runtime does not degrade as plans get longer. 
As the lower bound directing $A^*$-BFS
becomes less informed due to paths being removed from $G_k$, the more likely is BFS to devolve into Dijkstra's algorithm with
the added overheads of more complex data-structures and lower bound evaluations.

These results\footnote{The evaluation data along with \textsc{Jupyter} notebooks calculating these and other statistics can be found on the following
	link: \url{https://bit.ly/3BO07H2}.} provide compelling evidence of the practical feasibility of our Lazy PRM algorithm (Figure~\ref{fig:revised_decomposition}). The probability 
of producing a valid smooth plan $\rho$ under $1$ second is very high, on the benchmark considered here.

\section{Related Work}
\label{related_work}
Existing work on kinodynamic motion planning typically relies on discretizing trajectories
to check validity of plans~\cite{bohlin:lazy_prm:00,bhatia:10:icra,schulman:14:motion}, or require a triangulation
of $F$ into convex pieces~\cite{plaku:10:icra, bhatia:10:icra, shoukry:17:cdc}. A recent exception to this 
is the work of Zhang et al.~\cite{zhang:icaps:21}, who propose the
notion of Kinodynamic Networks, geometric graphs (section~\ref{probabilistic_roadmaps}) where $\Lambda$ are Cubic B\'{e}zier curves. 
A curve
is admitted into the definition of $\Lambda$ whenever its endpoints are vertices in the network and the convex hull of its control points does not
overlap any $O_i$. 
At the time of writing this we are not aware of published analysis showing this approach
to be AO or PC. The Lazy PRM algorithm in Figure~\ref{fig:revised_decomposition} is both AO and PC, and this follows directly from the 
theory developed for probabilistic roadmaps~\cite{karaman:11:ijrr,solovey:20:icra,solovey:20:prm}.

Conceptually close but methodologically distant to this paper is the recent work
of Ortiz-Haro et al.~\cite{ortiz-haro:conflict-driven} based on the Logic Geometric Programming (LGP) framework
of M. Toussaint~\cite{toussaint:15:ijcai}. Ortiz-Haro et al. propose to compile  
via polynomial-time reduction instances of Problem~\eqref{min_cost_path} into a discrete deterministic 
planning problem~\cite[Section 2.4]{lavalle:06}. 
Their
approach relies on very scalable methods for the latter class of optimal discrete control problems.
Ortiz-Haro et al. do not provide near-optimality guarantees, and the
overheads of representing the constraints of Problem~\eqref{min_cost_path} 
as STRIPS operators
are substantial

Lastly, the recent results showing Branch-and-Cut and other advanced techniques for Integer programming to 
be a highly performing approach to Multi-Agent Path Planning problems~\cite{lam:19:bcp}
were a direct inspiration to us. 
As it has long been recognized in the literature~\cite{lavalle:grid_prm,janson:ijrr}, the gap between 
so-called ``grid-based'' and sampling-based motion planning ``just'' lies in the provenance
and properties of the underlying geometric graphs both methodologies use.

\section{Future Work}
\label{future_work}
We look forward to study the relation between algorithms for Problem~\eqref{learning_problem} 
like PRM and the Dantzig-Wolfe decomposition for linear programming~\cite{wolsey:ip,desaulniers:05:cg}, developing 
theory to analyse possible connections between state-of-the-art motion planning algorithms and
classic results in optimization. We are also interested in further pursuing the research into the synthesis of motion plans that satisfy
LTL specifications initiated in the classic work by Bhatia et al.~\cite{bhatia:10:icra}, and to explore the potential of motion planning algorithms using 
\textsc{Polytrace} into model checking for hybrid control systems~\cite{doyen:18:model_checking_hybrid_systems}. Our approach supports
a very general class of geometric differential constraints of arbitrary order, those that are expressed as systems of polynomial equations.
We look forward to investigate the behaviour of the branch-and-cut planning algorithm proposed in this paper over more varied sets of constraints
and uncertainty~\cite{bekris:07:icra}.



%

\section*{Acknowledgment}

The authors would like to thank the Australian Government through Trusted Autonomous Systems,
a Defence Cooperative Research Centre under the Next Generation Technologies Fund.



\bibliographystyle{IEEEtran}
\bibliography{IEEEabrv,crossref,references}

\end{document}